\documentclass[letterpaper]{article} 
\usepackage{aaai2027}  
\usepackage[hyphens]{url}  
\usepackage{graphicx} 
\usepackage{natbib}  
\usepackage{caption} 
\usepackage{algorithm}
\usepackage{algorithmic}

\nocopyright 
\usepackage{newfloat}
\usepackage{listings}
\DeclareCaptionStyle{ruled}{labelfont=normalfont,labelsep=colon,strut=off} 
\floatstyle{ruled}
\newfloat{listing}{tb}{lst}{}
\floatname{listing}{Listing}

\usepackage{amsmath,amssymb,amsfonts}
\usepackage{booktabs}
\usepackage{array}
\usepackage{multirow}
\usepackage{bm}
\usepackage{tabularx}
\usepackage{makecell}   

\title{CoDe-SSM: Context-Detail Decoupled State Space Model for Efficient UHD Image Restoration}
\author{
    Jiaxu Su\textsuperscript{\rm 1}\equalcontrib,
    Zhijian Wu\textsuperscript{\rm 2}\equalcontrib,
    Jun Li\textsuperscript{\rm 1}\corresponding,
    Bo Zhang\textsuperscript{\rm 1},
    Yefeng Zheng\textsuperscript{\rm 2}\corresponding
}
\affiliations{
    \textsuperscript{\rm 1}School of Computer and Electronic Information, Nanjing Normal University, China\\
    \textsuperscript{\rm 2}Medical Artificial Intelligence Laboratory, Westlake University, China\\

    \{252202037, lijuncst, zhangbo\}@njnu.edu.cn, \{wuzhijian, zhengyefeng\}@westlake.edu.cn

}

\begin{document}
	\maketitle
	\begin{abstract}
		Ultra-high-definition (UHD) image restoration must balance the aggregation of spatially recurring degradation cues with the preservation of localized image structures. Compact aggregation can reduce redundant processing but may attenuate edges, textures, and other fine structures. Existing approaches manage UHD restoration cost through downsampling, window partitioning, or cluster-based token reduction; yet many of them do not explicitly retain information that is poorly represented by shared aggregation. In this study, we propose a Context-Detail Decoupled State Space Model (CoDe-SSM) for UHD restoration, which processes aggregated context and clustering residuals in separate pathways. The context modeling pathway, implemented by the Global Cluster Scan Module (GCSM), aggregates features into $K$ input-dependent cluster centers and applies selective SSM reasoning over the resulting fixed-order sequence, enabling cross-region context sharing while decoupling computational cost from spatial resolution. The detail recovery pathway, implemented by the Local High-Frequency Module (LHFM), processes the clustering residual with an input-derived high-frequency mask and a sparse mixture of convolutional experts. Extensive experiments on five UHD benchmarks and five degradation types demonstrate that our explicit context-detail decoupling strategy yields substantial gains in restoration quality while maintaining desirable efficiency.
	\end{abstract}

	\section{Introduction}
		Ultra-high-definition (UHD, $3840\times2160$) image restoration must simultaneously eliminate complex degradation components while preserving local details. This practical trade-off becomes acute at UHD resolution, where the pixel count makes both restoration quality and computational cost challenging.
		
		Image degradations such as uniform haze or globally consistent blur exhibit cross-region regularities; modeling them independently at every location can be redundant. At the same time, UHD images contain rich local structures, e.g., sharp edges and fine textures, whose fidelity is sensitive to compression. This tension suggests a natural-design principle: route shareable degradation context and unshareable structural details through separate processing pathways.
	    \begin{figure}[t]
			\centering
			\includegraphics[width=\columnwidth]{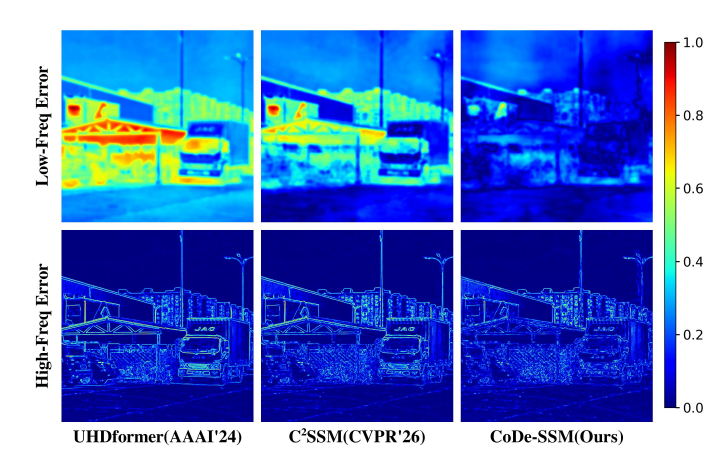}
			\caption{Illustrative comparison of low-frequency and high-frequency reconstruction errors. As clearly observed, our CoDe-SSM explicitly decouples global context and local details by combining prototype clustering and residual recovery, yielding significantly lower errors in both bands than UHDformer and C$^2$SSM in the displayed example.}
			\label{fig:lowhigh}
		\end{figure}
		Restoring UHD images at resolutions of 8 megapixels or higher poses a substantial computational challenge. CNNs and lightweight architectures can be efficient, but their receptive fields may limit global consistency modeling. Transformers~\cite{uhdformer,em:uformer,em:restormer} capture global dependencies but typically incur high quadratic memory and computational costs. State Space Models (SSMs)~\cite{mamba,vmamba,mambair,mambairv2} provide linear-complexity sequence reasoning; however, at UHD resolutions even linear scanning over the full pixel set can be demanding. To manage this cost, recent approaches adopt spatial compression techniques, such as downsampling~\cite{uhdfour} and patch partitioning~\cite{uhdformer}, or cluster-centric token aggregation~\cite{em:c2ssm}. However, as illustrated in Fig.~\ref{fig:lowhigh}, these methods apply uniform compression without explicitly separating components that can be shared from those that are poorly represented by aggregation. As a result, high-frequency textures may be attenuated. Prior work has not explicitly separated restoration information into aggregated context and a complementary detail pathway.

		To this end, we propose a Context-Detail Decoupled State Space Model (CoDe-SSM), which decouples the UHD restoration task into two dedicated pathways: a Global Cluster Scan Module (GCSM) that performs SSM reasoning over learned restoration prototypes for shared context correction, and a Local High-Frequency Module (LHFM) that recovers structural details from the clustering residual via operator-guided filtering and sparse expert routing. Experiments show that this decoupling strategy yields consistent improvements in restoration quality while preserving computational efficiency. In summary, our main contributions are threefold:
	\begin{itemize}
			\item We formulate the context-detail trade-off in UHD restoration, where recurring degradation patterns can benefit from aggregation while localized structures may be poorly represented by shared compression.
			\item We propose CoDe-SSM, a framework that decouples global context
			modeling from local detail preservation via two complementary
			pathways: a Global Cluster Scan Module (GCSM) for prototype-level SSM
			reasoning over compact cluster centers, and a Local High-Frequency
			Module (LHFM) for high-frequency-guided sparse expert routing.
			\item Extensive experiments demonstrate the effectiveness of CoDe-SSM across diverse benchmarks and various degradation types.
	\end{itemize}

	\section{Related Work}
	\subsection{UHD Image Restoration}
	UHD image restoration seeks to correct diverse degradations while preserving fine details. A central challenge is that full-resolution joint processing is computationally demanding, and compression strategies risk attenuating localized structures. Fourier-based methods such as UHDFour~\cite{uhdfour}, FFTformer~\cite{em:fftformer}, and Fourmer~\cite{em:fourmer} leverage frequency-domain processing for global modeling, yet their spectral transformations can introduce artifacts that attenuate fine spatial details. Transformer-based UHDformer~\cite{uhdformer} uses self-attention for context modeling. UHD-processer~\cite{uhdprocesser} leverages frequency-domain progressive learning within a VAE framework for all-in-one restoration, while UHDDIP~\cite{em:uhddip} introduces task-specific spatial abstraction, trading off fine-grained details against computational cost. DreamUHD~\cite{dreamuhd} adopts a frequency-enhanced VAE, integrating Fourier-based lightweight modules with wavelet adapters for high-frequency detail recovery. These methods largely approach UHD restoration as a computational-cost problem rather than an information-decoupling problem.
	\subsection{State Space Models for Vision}
	State Space Models (SSMs)~\cite{mamba} have been adapted for vision via 2D cross-scan paths~\cite{vmamba} and restoration backbones~\cite{mambair,mambairv2}. At UHD resolutions, however, even linear-complexity SSM scanning becomes computationally demanding. Recent efficient SSM designs, including EVSSM~\cite{em:evssm} for deblurring and MaIR~\cite{em:mair} for locality-preserving restoration, also face scalability challenges at full UHD scales. Methods like Wave-Mamba~\cite{wavemamba} and C\textsuperscript{2}SSM~\cite{em:c2ssm} reduce sequence length via frequency decomposition or cluster-centric token aggregation, but they apply a uniform compressed representation without explicit detail preservation.

	\section{Method}

	\subsection{Context-Detail Trade-off in UHD Restoration}
\begin{figure*}[tb]
		\centering
		\includegraphics[width=\textwidth]{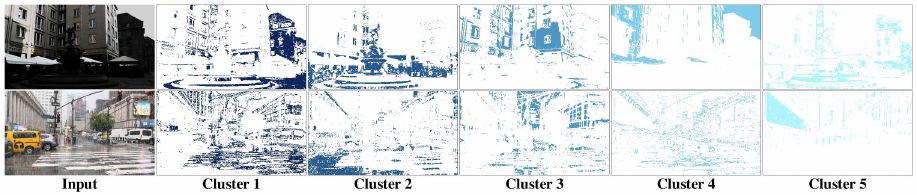}
		\caption{Visualization of representative cluster assignments. For clarity, we display 5 selected prototypes out of the $K$ used by the model. The network softly routes pixels with analogous degradation signatures into shared restoration prototypes, achieving semantic-aware context grouping by assigning structurally distinct regions to different prototypes.}
		\label{fig:cluster}
	\end{figure*}

\begin{figure}[tb]
		\centering
		\includegraphics[width=\columnwidth]{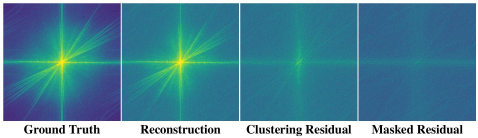}
		\caption{2D Fourier spectra of the cluster reconstruction and clustering residual. The cluster reconstruction predominantly captures low-frequency components, while the unmasked residual mixes discarded high-frequency details with low-frequency quantization errors. Our high-frequency masking mechanism effectively suppresses the low-frequency errors, yielding the isolated high-frequency details. This confirms that prototype aggregation discards fine-grained details.}
		\label{fig:error-down}
	\end{figure}

	As aforementioned, UHD restoration exhibits a context-detail trade-off: globally similar degradations benefit from shared correction, yet localized structures resist compression. CoDe-SSM addresses this through two observations:

	\begin{itemize}
		\item UHD images contain substantial pixel redundancy: cluster centers induced by learnable prototypes through soft clustering of full-image pixels can capture globally shared low-frequency context (refer to Fig.~\ref{fig:cluster}), enabling cross-region correction with reduced computation. However, high-frequency details are largely discarded during shared correction, as illustrated in Fig.~\ref{fig:error-down}.
		\item The clustering residual isolates the high-frequency details that are not addressed through shared correction. Given the heterogeneity of these unshareable details, their recovery can benefit from adaptive processing.
	\end{itemize}
	
	\begin{figure*}[tb]
		\centering
		\includegraphics[width=\textwidth]{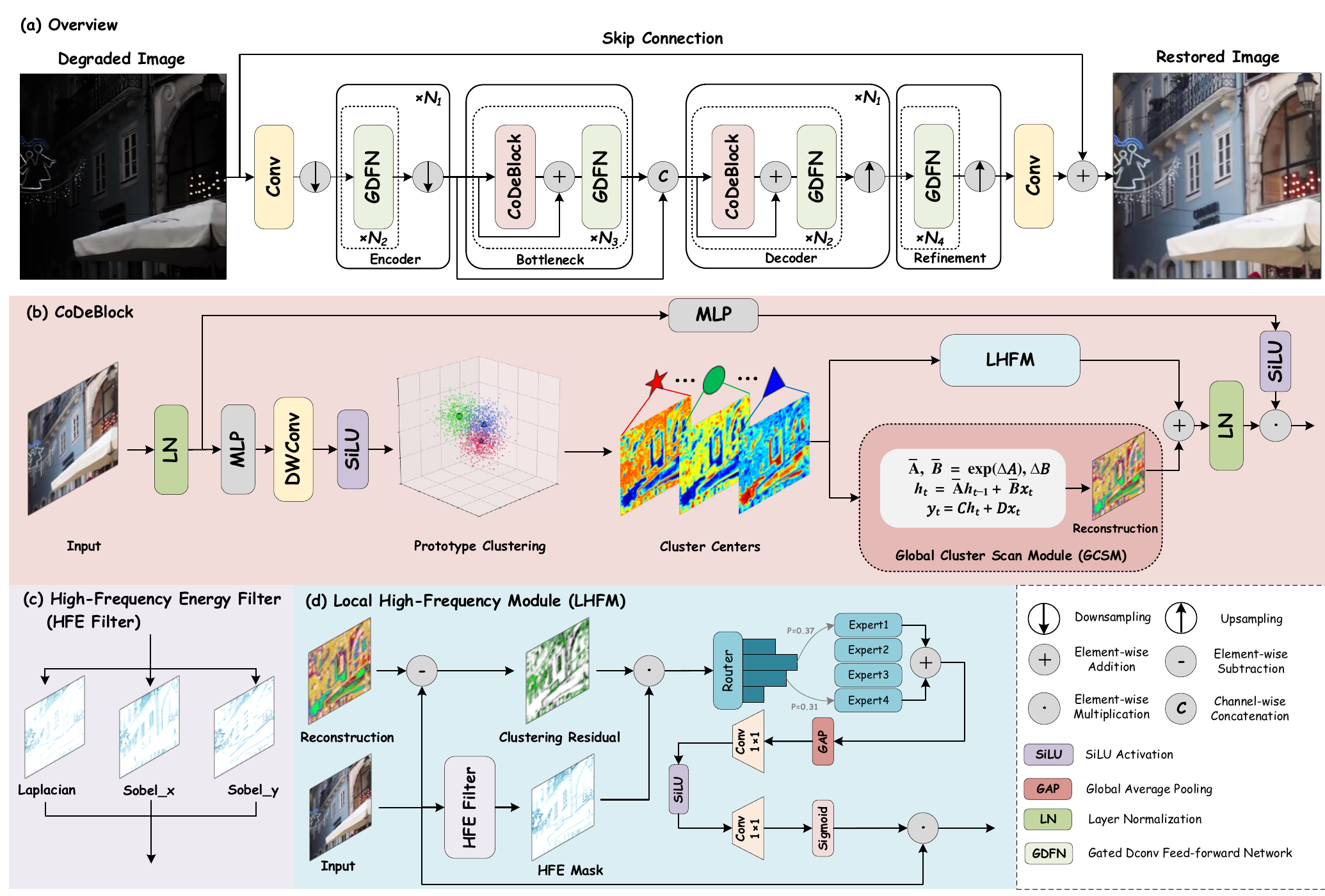}
		\caption{Architecture of CoDe-SSM. (a) The overall framework is an asymmetric U-Net with CoDeBlocks deployed at each encoder-decoder level. (b) The CoDeBlock clusters input features into $K$ centers, processes them through dual-branch GCSM along with LHFM pathways, and fuses the outputs via a GDFN. (c) The High-Frequency Energy (HFE) Filter extracts high-frequency maps from a single-channel grayscale reference using Laplacian and Sobel operators. (d) The Local High-Frequency Module (LHFM) applies an HFE mask to isolate structural details from the clustering residual, routes them through $E=4$ sparse convolutional experts with distinct receptive fields, and modulates the aggregated output via a channel-wise attention.}
		\label{fig:overview}
	\end{figure*}
	

	\subsection{Overall Architecture}
	As illustrated in Fig.~\ref{fig:overview}(a), CoDe-SSM adopts an asymmetric U-Net architecture. Given an input image $\mathbf{X} \in \mathbb{R}^{B \times 3 \times H \times W}$, a $3\times3$ convolution projects it into $C$ channels. The encoder comprises $N_1$ downsampling stages with $2\times2$ strided convolutions, progressively expanding channels from $C$ to $2C$ and $4C$. The decoder performs upsampling via pixel shuffling with skip connections. A learnable residual gate modulates the predicted restoration:
	\begin{equation}
		\hat{\mathbf{X}} = \mathbf{X} + \sigma(\alpha) \cdot f_\theta(\mathbf{X}),
		\label{eq:global_res}
	\end{equation}
	where $\sigma$ is the sigmoid function, $f_\theta(\mathbf{X})$ denotes the network-predicted residual, and $\alpha$ is a learnable scalar.
	
	Each U-Net level contains several CoDeBlocks. Let the input tokens at the current level be $\mathbf{Z} \in \mathbb{R}^{B \times N \times C}$, where $N = H_l W_l$ and $H_l, W_l$ denote the feature height and width. After Layer Normalization and a linear projection to $2C$ channels, the tokens are split into a feature branch $\mathbf{F} = \{\mathbf{f}_i\}_{i=1}^{N}$ and a gate branch $\mathbf{G} = \{\mathbf{g}_i\}_{i=1}^{N}$, both in $\mathbb{R}^{B \times N \times C}$.
	
	Before entering the dual pathways, we perform soft prototype clustering on the feature branch to separate shareable context from unshareable details. Specifically, we compute value features $\mathbf{V} = \{\mathbf{v}_i\}_{i=1}^{N}$ via a $1\times1$ convolution over $\mathbf{F}$, and softly assign them to $K$ learnable prototype anchors $\mathbf{P} = \{\mathbf{p}_k\}_{k=1}^{K} \in \mathbb{R}^{K \times C}$. The prototype anchors $\mathbf{P}$ are learnable parameters that define the clustering space, whereas the aggregated vectors $\mathbf{m}_k$ introduced below are input-dependent cluster centers. After $\ell_2$-normalization, we have:
	\begin{equation}
		\bar{\mathbf{v}}_i = \frac{\mathbf{v}_i}{\|\mathbf{v}_i\|_2 + \epsilon}, \qquad
		\bar{\mathbf{p}}_k = \frac{\mathbf{p}_k}{\|\mathbf{p}_k\|_2 + \epsilon},
		\label{eq:normalize}
	\end{equation}
	where $\epsilon$ is a small constant for numerical stability. The soft assignment is computed as
	\begin{equation}
		A_{ik} = \operatorname{softmax}_k\!\left( \frac{\langle \bar{\mathbf{v}}_i, \bar{\mathbf{p}}_k \rangle}{\tau} \right), \quad \tau = e^{\theta_\tau},
		\label{eq:soft_assign}
	\end{equation}
	where $A_{ik}$ is a scalar assignment probability and $\tau$ is a learnable temperature. Each prototype anchor aggregates its assigned tokens into an input-dependent cluster center:
	\begin{equation}
		\mathbf{m}_k = \frac{\sum_{i=1}^{N} A_{ik} \, \mathbf{v}_i}{\sum_{i=1}^{N} A_{ik} + \epsilon}.
		\label{eq:cluster_agg}
	\end{equation}
	The corresponding cluster reconstruction and clustering residual are
	\begin{align}
		\hat{\mathbf{v}}_i &= \sum_{k=1}^{K} A_{ik} \, \mathbf{m}_k, \label{eq:cluster_recon} \\
		\mathbf{R}_i &= \mathbf{v}_i - \hat{\mathbf{v}}_i. \label{eq:residual}
	\end{align}
	Thus, the clustering front-end produces three complementary outputs: the compact cluster centers $\{\mathbf{m}_k\}_{k=1}^{K}$, the soft assignment matrix $\mathbf{A}$ with entries $A_{ik}$, and the clustering residual $\mathbf{R}$.
	
	The cluster centers are passed to the context pathway, which refines them and maps them back to obtain $\mathbf{F}_{\text{global}}$. The residual is passed to the detail pathway, which recovers authentic high-frequency details to obtain $\mathbf{F}_{\text{local}}$. The two pathways are then fused and gated:
	\begin{align}
		\mathbf{F}_{\text{out}} &= \mathbf{F}_{\text{global}} + \mathbf{F}_{\text{local}}, \\
		\mathbf{F}'_{\text{out}} &= \operatorname{LN}(\mathbf{F}_{\text{out}}) \odot \operatorname{SiLU}(\mathbf{G}),
		\label{eq:fusion}
	\end{align}
	where $\operatorname{LN}$ denotes Layer Normalization, $\operatorname{SiLU}(x)=x\sigma(x)$, and $\odot$ is element-wise multiplication. The fused tokens are linearly projected and added back to $\mathbf{Z}$. After this fusion step, the resulting features are further processed by a GDFN~\cite{em:restormer}. All operations are performed independently at each U-Net level.

	\subsection{Context Modeling via Cluster Prototypes}
	The input-dependent cluster centers $\{\mathbf{m}_k\}_{k=1}^{K}$ summarize feature patterns captured by the soft assignments. The Global Cluster Scan Module (GCSM) therefore performs selective SSM reasoning~\cite{mamba} over this compact center sequence:
	\begin{equation}
		\{\mathbf{m}'_k\}_{k=1}^{K}
		=
		\operatorname{SSM}\!\big( \{\mathbf{m}_k\}_{k=1}^{K}; \theta_{\text{ssm}} \big).
		\label{eq:ssm_scan}
	\end{equation}
	Since $K \ll N$, this models cross-region degradation dependencies at a cost decoupled from the original spatial resolution. The refined centers are then mapped back to the pixel space through the soft assignment matrix:
	\begin{equation}
		\mathbf{F}_{\text{global}, i}
		=
		\sum_{k=1}^{K} A_{ik} \, \mathbf{m}'_k.
		\label{eq:broadcast}
	\end{equation}
	Because the assignment is image-global, distant pixels with similar degradation signatures can share refined prototypes, eliminating artificial spatial boundaries. As shown in Fig.~\ref{fig:cluster}, this mechanism groups semantically similar degradation regions into shared prototypes while routing structurally distinct regions separately.
	
	\subsection{Detail Recovery from the Residual}
	Prototype clustering yields a compact shared representation, though its reconstruction can attenuate localized detail. The clustering residual $\mathbf{R}$ retains the information not represented by that reconstruction (Fig.~\ref{fig:error-down}). However, this residual may include high-frequency signal as well as artifacts from the reconstruction step, making direct restoration suboptimal. The Local High-Frequency Module (LHFM) therefore applies an input-derived high-frequency mask to the residual and uses a sparse mixture of convolutional experts for adaptive reconstruction.
	
\begin{figure}[tb]
		\centering
		\includegraphics[width=\columnwidth]{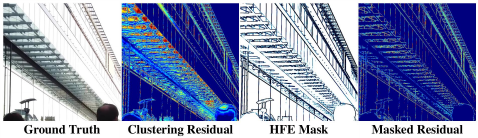}
		\caption{Visualization of Clustering residual and HFE Mask. The raw residual captures textures lost by prototype aggregation. The HFE mask suppresses flat-region artifacts, yielding a gated residual that faithfully isolates authentic high-frequency details.}
		\label{fig:error-up}
	\end{figure}

	To isolate details from the artifact-contaminated residual, we develop a High-Frequency Energy (HFE) Filter, as illustrated in Fig.~\ref{fig:overview}(c). Instead of relying on learnable parameters that might overfit to noise, the HFE Filter employs fixed Laplacian and Sobel operators on a single-channel reference map $\mathbf{Y}$, obtained by converting the original input image $\mathbf{X}$ to grayscale and resizing it to the current feature resolution $(H_l, W_l)$. The high-frequency energy map $\mathbf{E}_{\mathrm{hf}}$ is
	\begin{equation}
		\mathbf{E}_{\mathrm{hf}} = \frac{|\mathbf{L} \circledast \mathbf{Y}| + \sqrt{(\mathbf{S}_{x} \circledast \mathbf{Y})^{2} + (\mathbf{S}_{y} \circledast \mathbf{Y})^{2}}}{\mu_E + \epsilon},
		\label{eq:hf_energy}
	\end{equation}
	where $\mathbf{L}$ is the Laplacian kernel, $\mathbf{S}_{x}$ and $\mathbf{S}_{y}$ are the horizontal and vertical Sobel kernels, $\circledast$ denotes 2D convolution, and $\mu_E$ is the spatial mean of the numerator. This normalizes $\mathbf{E}_{\mathrm{hf}}$ to unit mean. A soft mask with learnable $\beta$ (initialized to $1.0$) gates the residual:
	\begin{equation}
		\mathbf{M} = \sigma\!\left(\beta(\mathbf{E}_{\mathrm{hf}} - 1)\right), \qquad
		\tilde{\mathbf{R}} = \mathbf{R} \odot \mathbf{M},
		\label{eq:refined_residual}
	\end{equation}
	where $\mathbf{M}$ is broadcast across feature channels. This suppresses artifacts in flat regions while preserving authentic high-frequency components. Fig.~\ref{fig:error-up} visualizes this process.
	
	UHD images contain highly diverse detail patterns, such as sharp contours, fine textures, and thin rain streaks, which are difficult to model with a single convolutional kernel. Therefore, we process the refined residual $\tilde{\mathbf{R}}$ with a sparse MoE module, as shown in Fig.~\ref{fig:overview}(d). For each spatial token $\tilde{\mathbf{r}}_i$, a lightweight router predicts expert probabilities $\pi_{i,e}$, from which the top-$2$ experts out of $N_e=4$ are selected and renormalized as gating weights $w_{i,e}$. The four experts are instantiated as depthwise separable convolutions with distinct kernel configurations: standard $3\times3$, dilated $3\times3$ with dilation rate $2$, cross-shaped $1\times5$ and $5\times1$, and dense $5\times5$, each followed by SiLU. The weighted expert outputs are aggregated and modulated by a channel-wise attention gate~\cite{em:senet} to produce $\mathbf{F}_{\text{local}}$.
	
	To prevent expert collapse, we employ a load-balancing loss~\cite{em:moe}. Let $\mathcal{T}_i$ denote the selected top-$2$ expert set for token $i$, and let $N$ be the number of tokens at the current level. We compute
	\begin{equation}
		f_e = \frac{1}{N}\sum_{i=1}^{N} \mathbf{1}[e \in \mathcal{T}_i], \qquad
		\bar{\pi}_e = \frac{1}{N}\sum_{i=1}^{N} \pi_{i,e},
		\label{eq:balance_stats}
	\end{equation}
	where $f_e$ is the fraction of tokens dispatched to expert $e$, and $\bar{\pi}_e$ is the mean router probability for expert $e$. The load-balancing loss is
	\begin{equation}
		\mathcal{L}_{\text{bal}} = N_e \sum_{e=1}^{N_e} f_e \, \bar{\pi}_e.
		\label{eq:balance}
	\end{equation}
	
	By exclusively targeting the clustering residual discarded by prototype aggregation, LHFM complements the context pathway without competing for representational capacity.

	\section{Experiments}
	\subsection{Experimental Setup}
	\textbf{Datasets.} We evaluate on five UHD benchmarks spanning five degradation types, summarized in Table~\ref{tab:datasets}.
	
	\noindent\textbf{Training.} We implement CoDe-SSM in PyTorch and train all models on 4 NVIDIA RTX 3090 GPUs. We randomly crop the full-resolution 4K images to a resolution of $768 \times 768$ as the input, with the batch size set to 4. For all UHD restoration tasks, we train for 200K iterations. To augment the training data, random horizontal and vertical flips are applied to the input images. Our method consists of an encoder-decoder with $N_1 = 3$ levels, where both the encoder and decoder share the same block structure: $N_2 = [2, 4, 4]$. The bottleneck and refinement contain $N_3 = N_4 = 4$ blocks, with a basic embedding dimension of 32. We adopt the AdamW optimizer with an initial learning rate of $5 \times 10^{-4}$, weight decay of $10^{-3}$, and cosine annealing. The cluster numbers are configured as $[16, 24, 32]$. To optimize the network, we utilize the $\ell_1$ loss and the FFT loss in the RGB color space as the basic reconstruction losses. Additionally, to prevent expert collapse in the MoE-based LHFM, we incorporate a load-balancing loss
	$\mathcal{L}_{\text{bal}}$ as defined in Eq.~\eqref{eq:balance}.
	
	\noindent\textbf{Evaluation.}  We report Structural Similarity Index Measure (SSIM)~\cite{em:ssim} and  Peak Signal-to-Noise Ratio (PSNR)~\cite{em:psnr} for datasets with ground-truth images. We additionally report Learned Perceptual Image Patch Similarity (LPIPS)~\cite{em:lpips} for perceptual quality assessment. We also compare parameter counts and floating-point operations (FLOPs) to evaluate model efficiency.

	\subsection{Results and Discussion}
	Table~\ref{tab:lol4k} reports quantitative results on UHD-LOL4K. CoDe-SSM achieves 42.24 dB PSNR, surpassing the previous best C\textsuperscript{2}SSM by 2.63 dB. On UHD-Haze, CoDe-SSM attains 27.09 dB PSNR, a 3.01 dB gain over C\textsuperscript{2}SSM (Table~\ref{tab:haze}). For spatially sparse degradations, CoDe-SSM reaches 42.24 dB on UHD-Rain and 43.00 dB on UHD-Snow, exceeding leading methods by 2.07 dB and 0.55 dB, respectively (Tables~\ref{tab:rain} and \ref{tab:snow}). On UHD-Blur, CoDe-SSM achieves 31.75 dB PSNR, outperforming C\textsuperscript{2}SSM by 0.22 dB and ERR by 2.03 dB (Table~\ref{tab:deblur}).
	
\begin{figure*}[t]
		\centering
		\includegraphics[width=\textwidth]{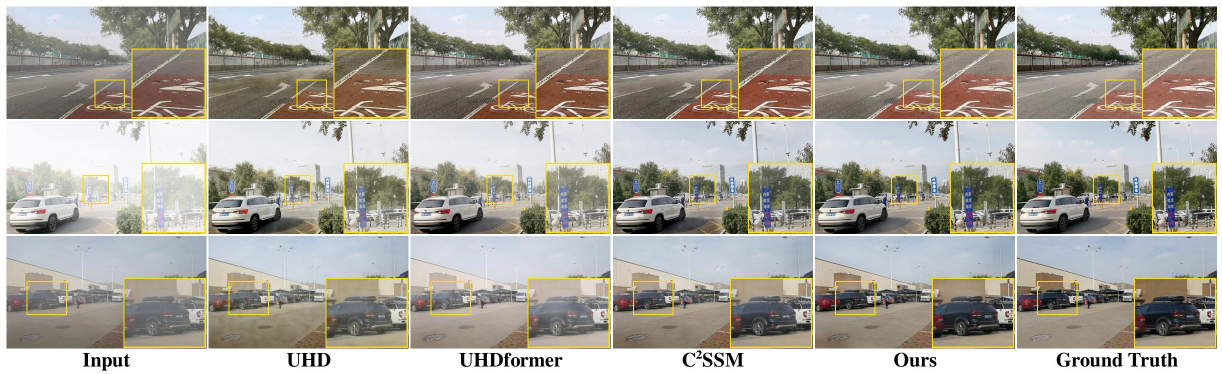}
		\caption{Qualitative comparison of our CoDe-SSM and state-of-the-art methods on the UHD-Haze dataset.}
		\label{fig:qual_haze}
	\end{figure*}

\begin{figure*}[t]
		\centering
		\includegraphics[width=\textwidth]{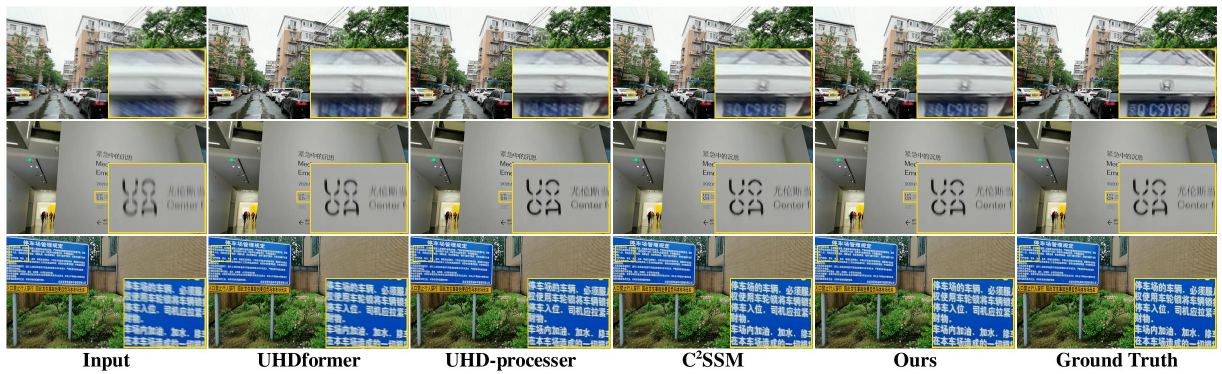}
		\caption{Qualitative comparison of our CoDe-SSM and state-of-the-art methods on the UHD-Blur dataset.}
		\label{fig:qual_deblur}
	\end{figure*}

	Beyond distortion metrics, we assess perceptual quality through LPIPS on the UHD-Rain dataset (Table~\ref{tab:lpips}). CoDe-SSM attains an LPIPS of 0.019, substantially lower than the next best method UHDDIP (0.030) and far below earlier methods Uformer and Restormer (0.460 and 0.478). This pronounced perceptual advantage indicates that the high-frequency detail preserved by the decoupled LHFM pathway contributes directly to feature-level similarity, beyond what PSNR gains alone would capture. Visual comparisons are shown in Figs.~\ref{fig:qual_haze} and \ref{fig:qual_deblur}.
	
	These results validate the complementary roles of the two pathways: GCSM-driven context modeling dominates on spatially consistent degradations (low-light, haze), while LHFM-driven detail recovery proves essential for sparse, irregular scenarios (rain, snow).
	\begin{table}[tb]
		\centering
		\small
		\setlength{\tabcolsep}{15pt}
		\begin{tabularx}{\columnwidth}{@{} >{\raggedright\arraybackslash}X | c | c @{}}
			\toprule
			\textbf{Dataset} & \textbf{Train} & \textbf{Test} \\
			\midrule
			UHD-LOL4K~\cite{em:uhdlol4k}  & 5,999  & 2,100 \\
			UHD-Haze~\cite{uhdformer}   & 2,290  & 231 \\
			UHD-Blur~\cite{uhdformer}   & 1,964  & 300 \\
			UHD-Snow~\cite{em:uhddip}   & 3,000  & 200 \\
			UHD-Rain~\cite{em:uhddip}   & 3,000  & 200 \\
			\bottomrule
		\end{tabularx}
		\caption{Dataset statistics for various UHD image restoration tasks.}
		\label{tab:datasets}
	\end{table}

	\begin{table}[tb]
		\centering
		\small
		\setlength{\tabcolsep}{2.5pt}
		\begin{tabularx}{\columnwidth}{@{} >{\raggedright\arraybackslash}X | c | c c | c @{}}
			\toprule
			\textbf{Methods} & \textbf{Venue} & \textbf{PSNR} & \textbf{SSIM} & \textbf{Param.} \\
			\midrule
			NSEN~\cite{em:nsen}        & MM     & 29.49 & 0.980 & 2.67M \\
			UHDFour~\cite{uhdfour}     & ICLR   & 36.12 & 0.990 & 17.5M \\
			LLFormer~\cite{em:uhdlol4k}    & AAAI   & 37.33 & 0.988 & 24.5M \\
			UHDformer~\cite{uhdformer} & AAAI   & 36.28 & 0.989 & 0.34M \\
			Wave-Mamba~\cite{wavemamba}  & MM     & 37.43 & 0.990 & 1.25M \\
			D2Net~\cite{em:d2net}       & WACV   & 37.73 & 0.992 & 5.22M \\
			C\textsuperscript{2}SSM~\cite{em:c2ssm}   & CVPR   & 39.61 & 0.992 & 2.71M \\			
			\textbf{CoDe-SSM} & \textbf{Ours}        & \textbf{42.24} & \textbf{0.996} & 2.88M \\
			\bottomrule
		\end{tabularx}
		\caption{Quantitative comparison with state-of-the-art methods on the UHD-LOL4K dataset.}
		\label{tab:lol4k}
	\end{table}

	\begin{table}[tb]
		\centering
		\small
		\setlength{\tabcolsep}{0.9pt}
		\begin{tabularx}{\columnwidth}{@{} >{\raggedright\arraybackslash}X | c | c c | c @{}}
			\toprule
			\textbf{Methods} & \textbf{Venue} & \textbf{PSNR} & \textbf{SSIM} & \textbf{Param.} \\
			\midrule
			UHDformer~\cite{uhdformer} & AAAI  & 28.82 & 0.844 & 0.34M \\
			UHDDIP~\cite{em:uhddip} & TCSVT & 28.28 & 0.845 & 0.81M \\
			DreamUHD~\cite{dreamuhd} & AAAI  & 29.33 & 0.852 & 1.45M \\
			UHD-processer~\cite{uhdprocesser} & CVPR  & 29.43 & 0.855 & 1.60M \\
			ERR~\cite{em:err}           & CVPR  & 29.72 & 0.861 & 1.13M \\
			C\textsuperscript{2}SSM~\cite{em:c2ssm}   & CVPR   & 31.53 & 0.890 & 2.71M\\					
			\textbf{CoDe-SSM} & \textbf{Ours}              & \textbf{31.75} & \textbf{0.894} & 2.88M \\
			\bottomrule
		\end{tabularx}
		\caption{Quantitative comparison with state-of-the-art methods on the UHD-Blur dataset.}
		\label{tab:deblur}	
	\end{table}

	\begin{table}[tb]
		\centering
		\small
		\setlength{\tabcolsep}{0.9pt}
		\begin{tabularx}{\columnwidth}{@{} >{\raggedright\arraybackslash}X | c | c c | c @{}}
			\toprule
			\textbf{Methods} & \textbf{Venue} & \textbf{PSNR} & \textbf{SSIM} & \textbf{Param.} \\
			\midrule
			UHD~\cite{em:uhd} & ICCV  & 18.04 & 0.811 & 34.5M \\
			UHDformer~\cite{uhdformer} & AAAI  & 22.59 & 0.942 & 0.34M \\
			UHDDIP~\cite{em:uhddip} & TCSVT & 22.14 & 0.941 & 0.81M \\
			UHD-processer~\cite{uhdprocesser}   & CVPR  & 23.24 & 0.953 & 1.60M \\
			C\textsuperscript{2}SSM~\cite{em:c2ssm}   & CVPR   & 24.08 & 0.942 & 2.71M\\	
			\textbf{CoDe-SSM} & \textbf{Ours}        & \textbf{27.09} & \textbf{0.963} & 2.88M \\
			\bottomrule
		\end{tabularx}
		\caption{Quantitative comparison with state-of-the-art methods on the UHD-Haze dataset.}
		\label{tab:haze}
	\end{table}
	\begin{table}[tb]
		\centering
		\small
		\setlength{\tabcolsep}{0.9pt}
		\begin{tabularx}{\columnwidth}{@{} >{\raggedright\arraybackslash}X | c | c c | c @{}}
			\toprule
			\textbf{Methods} & \textbf{Venue} & \textbf{PSNR} & \textbf{SSIM} & \textbf{Param.} \\
			\midrule
			Uformer~\cite{em:uformer} & CVPR  & 19.49 & 0.716 & 50.9M \\
			Restormer~\cite{em:restormer} & CVPR  & 19.41 & 0.711 & 25.3M \\
			SFNet~\cite{em:sfnet} & ICLR  & 20.10 & 0.709 & 13.3M \\
			UHDformer~\cite{uhdformer} & AAAI  & 37.34 & 0.974 & 0.34M \\
			UHDDIP~\cite{em:uhddip} & TCSVT  & 40.17 & 0.982 & 0.81M \\	
			\textbf{CoDe-SSM} & \textbf{Ours}        & \textbf{42.24} & \textbf{0.989} & 2.88M \\
			\bottomrule
		\end{tabularx}
		\caption{Quantitative comparison with state-of-the-art methods on the UHD-Rain dataset.}
		\label{tab:rain}
	\end{table}
	
	\begin{table}[tb]
		\centering
		\small
		\setlength{\tabcolsep}{0.9pt}
		\begin{tabularx}{\columnwidth}{@{} >{\raggedright\arraybackslash}X | c | c c | c @{}}
			\toprule
			\textbf{Methods} & \textbf{Venue} & \textbf{PSNR} & \textbf{SSIM} & \textbf{Param.} \\
			\midrule
			Uformer~\cite{em:uformer} & CVPR  & 23.72 & 0.871 & 50.9M \\
			Restormer~\cite{em:restormer} & CVPR  & 24.14 & 0.869 & 25.3M \\
			SFNet~\cite{em:sfnet} & ICLR  & 23.64 & 0.846 & 13.3M \\
			UHDformer~\cite{uhdformer} & AAAI  & 36.61 & 0.988 & 0.34M \\
			UHDDIP~\cite{em:uhddip} & TCSVT & 41.56 & 0.990 & 0.81M \\
			C\textsuperscript{2}SSM~\cite{em:c2ssm}   & CVPR  & 42.45 & 0.990 & 2.71M\\		
			\textbf{CoDe-SSM} & \textbf{Ours}        & \textbf{43.00} & \textbf{0.992} & 2.88M \\
			\bottomrule
		\end{tabularx}
		\caption{Quantitative comparison with state-of-the-art methods on the UHD-Snow dataset.}
		\label{tab:snow}
	\end{table}

	\begin{table}[tb]
		\centering
		\small
		\setlength{\tabcolsep}{10pt}
		\begin{tabularx}{\columnwidth}{@{} >{\raggedright\arraybackslash}X | c | c @{}}
			\toprule
			\textbf{Methods} & \textbf{Venue} & \textbf{LPIPS}$\downarrow$ \\
			\midrule
			Uformer~\cite{em:uformer}                    & CVPR    & 0.460 \\
			Restormer~\cite{em:restormer}                & CVPR    & 0.478 \\
			SFNet~\cite{em:sfnet}                        & ICLR    & 0.477 \\
			UHDformer~\cite{uhdformer}                   & AAAI    & 0.055 \\
			UHDDIP~\cite{em:uhddip}                      & TCSVT  & 0.030 \\
			\textbf{CoDe-SSM}                     & \textbf{Ours}     & \textbf{0.019} \\
			\bottomrule
		\end{tabularx}
		\caption{LPIPS comparison with state-of-the-art methods on the UHD-Rain dataset.}
		\label{tab:lpips}
	\end{table}
	\begin{table}[t] 
		\centering
		\small
		\setlength{\tabcolsep}{5pt}
		\begin{tabularx}{\columnwidth}{@{} >{\raggedright\arraybackslash}X | c | c | c @{}}
			\toprule
			\textbf{Methods} & \textbf{Venue} & \textbf{Param.} & \textbf{FLOPs} \\
			\midrule
			D2Net~\cite{em:d2net}        & WACV    & 5.22M & 148.84G \\
			Wave-Mamba~\cite{wavemamba}  & MM      & 1.25M & 28.73G  \\
			UDR-Mixer~\cite{em:rain13k}  & TMM     & 4.90M & 52.57G  \\
			C\textsuperscript{2}SSM~\cite{em:c2ssm} & CVPR & 2.71M & 26.04G  \\
			\textbf{CoDe-SSM} & \textbf{Ours}        & 2.88M & 25.83G  \\
			\bottomrule 
		\end{tabularx}
		\caption{Efficiency comparison of different UHD methods. FLOPs are measured with an input size of $512 \times 512$.}
		\label{tab:efficiency}
	\end{table}

	\begin{table}[t]
		\centering
		\small
		\setlength{\tabcolsep}{7pt}
		\begin{tabularx}{\columnwidth}{@{}
				>{\raggedright\arraybackslash}X
				| c | c | c
				@{}}
			\toprule
			\textbf{Variant} & \textbf{PSNR}$\uparrow$ & \textbf{SSIM}$\uparrow$ & \textbf{Param.} \\
			\midrule
			(a) GCSM $\rightarrow$ ResBlock      & 39.18 & 0.991 & 2.91M \\
			(b) LHFM $\rightarrow$ FFN      & 41.26 & 0.993 & 3.12M \\
			(c) w/o GCSM                     & 37.49 & 0.986 & 2.54M \\
			(d) w/o LHFM                     & 40.51 & 0.992 & 2.83M \\
			(e) w/o HFE Filter               & 41.92 & 0.995 & 2.88M \\
            (f) w/o MoE               & 41.81 & 0.995 & 2.79M \\
			(g) Complete model (Ours)               & \textbf{42.24} & \textbf{0.996} & 2.88M \\
			\bottomrule
		\end{tabularx}
		\caption{Ablation study of proposed blocks on the UHD-LOL4K dataset.}
		\label{tab:ablation}
	\end{table}
	
	\begin{table}[t]
		\centering
		\small
		\setlength{\tabcolsep}{3pt}
		\begin{tabularx}{\columnwidth}{@{}
				l 
				| >{\centering\arraybackslash}X 
				| >{\centering\arraybackslash}X 
				| >{\centering\arraybackslash}X 
				| >{\centering\arraybackslash}X 
				@{}}
			\toprule
			\textbf{Metric} & [8,12,16] & [12,16,24] & [16,24,32] & [24,32,48] \\
			\midrule
			\textbf{PSNR}$\uparrow$  & 40.41 & 41.53 & \textbf{42.24} & 41.14 \\
			\textbf{SSIM}$\uparrow$  & 0.995 & 0.995 & \textbf{0.996} & 0.994 \\
			\textbf{Param.} & 2.86M & 2.87M & 2.88M & 2.89M \\
			\bottomrule
		\end{tabularx}
		\caption{Ablation study of the number of clusters $K$ on the UHD-LOL4K dataset.}
		\label{tab:ablation_k}
	\end{table}
	
	\begin{table}[tb]
		\centering
		\small
		\setlength{\tabcolsep}{12pt}
		\begin{tabularx}{\columnwidth}{@{} >{\raggedright\arraybackslash}X | c| c| c |c @{}}
			\toprule
			\textbf{Metric} & \textbf{$K\!=\!1$} & \textbf{$K\!=\!2$} & \textbf{$K\!=\!3$} & \textbf{$K\!=\!4$} \\
			\midrule
			\textbf{PSNR}$\uparrow$  
			& 41.94 
			& \textbf{42.24} 
			& 41.28 
			& 41.87 \\
			
			\textbf{SSIM}$\uparrow$  
			& 0.993 
			& \textbf{0.996} 
			& 0.993 
			& 0.992 \\
			
			\textbf{LPIPS}$\downarrow$ 
			& 0.0134 
			& \textbf{0.0121} 
			& 0.0145 
			& 0.0148 \\
			\bottomrule
		\end{tabularx}
		\caption{Ablation study on the number of activated experts (Top-K) in MoE on the UHD-LOL4K dataset.}
		\label{tab:ablation_topk}
	\end{table}
	
	\subsection{Computational Efficiency Analysis}

		Table~\ref{tab:efficiency} reports the empirical efficiency of CoDe-SSM compared with recent methods. CoDe-SSM achieves the lowest FLOPs among all compared methods with 2.88M parameters, confirming that decoupling context from details yields both high restoration quality and lower computational cost.

	For theoretical analysis, we analyze the per-block computational complexity. Let $N = H_l W_l$, $C$, and $K$ denote the spatial token count, feature dimension, and number of prototypes ($K \ll N$). GCSM performs soft clustering, prototype aggregation, SSM scanning over $K$ centers, and broadcast, yielding $\mathcal{O}(N K C + K C^{2})$. LHFM applies fixed-operator filtering and sparse MoE convolutions to the residual at $\mathcal{O}(N C)$. Given $K \ll N$, the total complexity per CoDeBlock is roughly $\mathcal{O}(N K C + K C^{2})$.

	\subsection{Ablation Studies}
	\noindent\textbf{Component Analysis.} Table~\ref{tab:ablation} reports the contribution of each component. Removing GCSM and LHFM reduces PSNR by 4.75\thinspace dB and 1.73\thinspace dB, respectively, indicating that both pathways contribute to restoration performance. Replacing GCSM with a ResBlock causes a 3.06\thinspace dB decline, while replacing LHFM with a standard FFN incurs a 0.98\thinspace dB degradation. Removing the MoE sub-module from LHFM drops PSNR by 0.43\thinspace dB, and removing the HFE structural gate costs 0.32\thinspace dB decline.
	
	\noindent\textbf{Cluster Count.} Table~\ref{tab:ablation_k} studies the effect of the cluster count $K$. PSNR follows an inverted-U pattern across the tested settings, peaking at 42.24\thinspace dB with $[16,24,32]$, compared with 40.41\thinspace dB at $[8,12,16]$ and 41.14\thinspace dB at $[24,32,48]$. Parameter counts vary by only 0.03M. This pattern is consistent with a trade-off between representation granularity and center compactness: too few clusters under-represent fine degradation patterns, while too many clusters dilute prototype compactness.
	
	\noindent\textbf{Expert Activation.} Table~\ref{tab:ablation_topk} studies the effect of Top-$K$ in the MoE module. Performance follows an inverted-U trend: PSNR, SSIM, and LPIPS all peak at $K\!=\!2$, reaching 42.24\thinspace dB, 0.996, and 0.0121, respectively. Reducing to $K\!=\!1$ degrades PSNR by 0.30\thinspace dB and increases LPIPS by 0.0013. Increasing to $K\!=\!3$ causes a sharper PSNR drop of 0.96\thinspace dB and a corresponding LPIPS rise to 0.0145. $K\!=\!4$ partially recovers to 41.87\thinspace dB but remains below $K\!=\!2$.
	
	\section{Conclusion}
	In this study, we proposed CoDe-SSM, a context-detail decoupled framework for efficient UHD image restoration. By routing shared degradation context through prototype-based SSM reasoning (GCSM) and recovering localized structures from the clustering residual via operator-guided sparse experts (LHFM), CoDe-SSM achieves strong performance across five benchmarks with only 2.88M parameters.
	A limitation is that the high-frequency energy gate in LHFM relies on global-mean normalization, which may amplify noise or suppress weak details under highly non-uniform illumination. Future work will explore illumination-aware local normalization for more robust detail recovery and extend CoDe-SSM toward unified, degradation-agnostic restoration.

	\bibliography{aaai2027}

\end{document}